\documentclass[11pt]{article}

\usepackage[utf8]{inputenc}
\usepackage[T1]{fontenc}
\usepackage{amsmath, amssymb}
\usepackage{graphicx}
\usepackage{hyperref}
\usepackage{geometry}
\usepackage{cite}
\usepackage{float}

\usepackage{algorithm}
\usepackage{algpseudocode}

\hypersetup{
  colorlinks=true,
  linkcolor=blue,
  citecolor=blue,
  urlcolor=blue
}

\title{Features-based embedding or Feature-grounding}
\author{Piotr Makarevich\thanks{Email: \texttt{peter.makarevich.wrk@gmail.com}}}
\date{\today}

\begin{document}

\maketitle

\begin{abstract}
  In everyday reasoning, when we think about a particular object, we associate it with a unique set of expected properties such as weight, size, or more abstract attributes like density or horsepower.
  These expectations are shaped by our prior knowledge and the conceptual categories we have formed through experience.
  This paper investigates how such knowledge-based structured thinking can be reproduced in deep learning models using features based embeddings.
  Specially, it introduces an specific approach to build feature-grounded embeddings, aiming to align shareable representations of operable dictionary with interpretable domain-specific conceptual features.
\end{abstract}

\section{Introduction}

  Pre-trained language models such as BERT~\cite{bert} have become foundational in modern natural language processing, owing to their ability to capture rich contextual representations from large-scale corpora.
  However, it is still unclear where and how extracted knowledge from training data is internally represented in the model, and how we can distribute this knowledge between structurally similar models.

  This work introduces a specific method for word-embedding initialization that encapsulates domain-specific knowledge into internal representations to construct feature-grounded embeddings.
  Such kind of embedding provides structured prior into internal weight landscape during LLM training. 

\section{Related Work}

The idea of augmenting token embeddings with structured information has been extensively studied in natural language processing (NLP), particularly in the context of \emph{positional encodings}, as introduced in Transformer-based architectures~\cite{vaswani2017attention}. Positional encodings inject information about the absolute or relative positions of tokens within a sequence, thereby compensating for the permutation-invariant nature of self-attention mechanisms.

Conventional positional encodings, whether fixed (e.g., sinusoidal) or learned, are typically \textit{additive} components applied to token embeddings. They are not grounded in external conceptual or semantic knowledge but rather served to encode \textit{sequence-level ordering information}.

By contrast, the approach proposed in this work represents a conceptual shift. Instead of encoding positional order, it introduces domain-specific, interpretable structure into the word-embedding space
 \emph{prior} to training, that allows to encode physical, functional or abstract semantic knowledge derived from an external source (e.g., a knowledge base or pretrained model).
This design introduces a continuous, geometry-aware transformation specific to each token, with training guided by a combination of losses in the projected space.
While both positional encodings and feature-grounding impose structural priors on embeddings, their objectives and mechanisms differ fundamentally, as summarized in Table~\ref{tab:pe_vs_rso}.

\begin{table}[h]
\centering
\caption{Comparison between Positional Encoding (PE) and feature-grounding (FE)}
\label{tab:pe_vs_rso}
\begin{tabular}{|p{3.5cm}|p{5.2cm}|p{5.2cm}|}
\hline
\textbf{Aspect} & \textbf{Positional Encoding (PE)} & \textbf{Rotated Saturation Operator (RSO)} \\
\hline
Purpose & Encode sequence order & Encode domain knowledge and structural priors \\
\hline
Inductive Bias & Sequence-aware modeling & Conceptually grounded embedding geometry \\
\hline
Loss Function & None explicitly; optimized indirectly & MSE + contrastive loss on feature space \\
\hline
Knowledge Source & Sequence structure & External ontologies, embeddings, or domain-specific knowledge \\
\hline
\end{tabular}
\end{table}

Moreover, whereas positional encodings operate uniformly across layers, the saturation operator acts at the embedding level during training and introduces \emph{localized, token-specific transformation} prior to any further network processing. This aligns the learned representations with structured priors in a way that is interpretable and potentially transferable across tasks or domains.

The proposed approach introduces a novel mechanism for embedding pretraining, grounded in domain-specific conceptual features and implemented through a structured, non-learnable projection operator. 

Formally, each token is associated with a vector of structured features, which are projected from the embedding space using a soft lower-triangular operator modulated by a rotational parameter dependent on token index. 
This operation yields a projected embedding space optimized to preserve both local feature fidelity (via reconstruction loss) and global relational geometry (via contrastive loss with min-max regularization).

Taken together, this method unifies insights from grounded representation learning, structured projection geometry, and contrastive objectives. 
It functions as an architecture-agnostic mechanism that improves the interpretability of word-embeddings and facilitates semantic interoperability across models.
As a result, it enables the word-embedding layer itself to serve as a carrier of high-level inductive priors, with potential applications in transfer learning, domain adaptation and representation auditing.

\newpage
\section{Method}

\subsection{The Idea:}

The central idea is: before training a large language model, conceptual information derived from domain ontologies, inject directly into the space representation of the embedder, which will shape the entire network during standard training.

Each embedding token (in the context of word embeddings) is implicitly associated with a predefined set of features that capture the expected properties or categorical roles of the corresponding lexical unit.
These features may reflect physical attributes (e.g., size, weight), functional characteristics (e.g., tool, transport), or abstract conceptual dimensions (e.g., animal, building), depending on the target domain.

\subsection{The approach:}

\textbf{Preparation for learning:}

  Each token of embedding dictionary can be denoted as \( t \in \mathcal{V} \), where \(\mathcal{V}\) is a finite vocabulary. The token \( t \) refers to a specific position (or index) of a word within this vocabulary.

  For each token \( t \in \mathcal{V} \), exists a set of domain knowledge facts, denoted as
  \[
  \mathbf{f}_t \in \mathbb{R}^k,
  \]
  
  where \(\mathbf{f}_t\) is a feature vector encoding prior knowledge about \(t\).
  
  These facts can represent:
  \begin{itemize}
      \item Quantitative physical measurements (e.g., weight, size),
      \item Strictly abstract categorical concepts (e.g., semantic categories, functional roles).
  \end{itemize}
  
  These facts for each token can be extracted from a data source (e.g., a large language model or a database) in the form of a knowledge-based feature vector.
  \vspace{1em}
  
  Formally, this can be summarized as:
  \[
  \begin{cases}
  t \in \mathcal{V}, & \text{token from vocabulary} \\
  \mathbf{f}_t \in \mathbb{R}^k, & \text{knowledge-based feature vector}
  \end{cases}
  \]

  Let the matrix of all token-level feature vectors be denoted as:
  \[
  X := 
  \begin{bmatrix}
  \mathbf{f}_{t_1}^\top \\
  \mathbf{f}_{t_2}^\top \\
  \vdots \\
  \mathbf{f}_{t_T}^\top
  \end{bmatrix}
  \in \mathbb{R}^{T \times k},
  \]

\vspace{1em}
\textbf{Learning Objectives:}

  During the embedding pretraining process, weights will be updated using a combination of two loss functions:
  \begin{itemize}
    \item \textbf{Reconstruction Loss:} Mean Squared Error (MSE), used to preserve the low-level feature structure.
    \item \textbf{Structural Loss:} Contrastive Loss with additional min/max distance penalties, used to shape relational geometry between tokens and work as regularizer term as well.
  \end{itemize}

\textbf{Saturation Operator:}

  To enable a more compact space representation and to prevent conflicts arising from similar features across different tokens, the technique introduces a specialized projection matrix referred to as the - \textbf{saturation operator}.
  
  This projection matrix is defined as a soft-triangular matrix, with non-zero elements (e.g. 0.55 below the main diagonal and 0.45 above it). 
  It serves to modulate the output of the embedding layer and can be interpreted as a non-learnable low-dimensional bottleneck for the training purpose.
  
  The matrix projects the embedding space into a lower-dimensional feature space, allowing calculation of the feature - \textbf{reconstruction loss}.
  To reduce conflicts caused by feature similarity and spatial proximity among token embeddings, the soft-triangular matrix is rotated according on each token`s position within the embedding dictionary. As a result, this process constructs a distinct embedding-to-feature output gate for each token.

  It can be written as
   \( R_z \in \mathbb{R}^{|\mathcal{V}| \times \mathbf{f}_t} \)
  as a soft lower-triangular projector matrix with rotational structure.

  The operator performs a structured transformation by applying a rotation around the central axis, controlled by a normalized angular parameter.

  Formally, define the normalized rotation angle \(\theta_t\) as:

    \[
    \theta_t = \frac{t}{|\mathcal{V}| + 1},
    \]

where \(|\mathcal{V}|\) is the size of the vocabulary, and |t| - token position. 

\vspace{1em}
The matrix \( R_z(\theta_t) \) satisfies the following properties:
\begin{itemize}
  \item approximate lower-triangularity,
  \item non-zero elements,
  \item for each index t, the rotation matrix is parameterized by the normalized rotation angle \(\theta_t\).
\end{itemize}

\vspace{1em}
The full token-specific saturation operator is defined as:

\[
\tilde{R}_t = R_z \cdot \mathcal{R}(\theta_t) \in \mathbb{R}^{d \times f}
\]

Embedding output is defined as:
\[
  \mathbf{e}_t \in \mathbb{R}^d
\]
where \( \mathbf{e}_t \) is the embedding representation of the token \( t \).

\vspace{1em}
\textbf{Saturation operator} \(\tilde{R}_t\) is applied to the embedding vector \( \mathbf{e}_t \in \mathbb{R}^d \) as follows:

\[
\tilde{\mathbf{e}}_t = \tilde{R}_t^\top \mathbf{e}_t \in \mathbb{R}^f
\]

Then
\[
\tilde{E} := \left[ \tilde{\mathbf{e}}_1 \ \tilde{\mathbf{e}}_2 \ \dots \ \tilde{\mathbf{e}}_T \right]^\top \in \mathbb{R}^{T \times f}
\] 
The resulting vector $\tilde{E}$ is denoted as the projected embeddings (or fuzzy embeddings).

\paragraph{Summary:}

\[
\mathbf{e}_i = \mathrm{emb}(t), \qquad \tilde{\mathbf{e}}_i = R_t \mathbf{e}_i
\]

\[
\mathcal{L} = 
\underbrace{\mathcal{L}_{\mathrm{MSE}}(\tilde{E}, X)}_{\text{Reconstruction Loss}} +   \underbrace{\mathcal{L}_{\mathrm{contrastive}}(\mathbf{e}_{t_i}, \mathbf{e}_{t_{j}})}_{\text{Min-max Distance Loss}}
\]

\begin{algorithm}[H]
  \caption{Saturation operator}
  \begin{algorithmic}[1]

    \State Compute normalized angle:
    \[
      \theta_t = \frac{t}{|V| + 1}, \quad \theta_t \in [0, 1]
    \]
    \Comment{where \(|V|\) denotes the vocabulary size, and \(t\) is the token position}
    
    \State Construct rotation matrix:
    \[
      \mathcal{R}(\theta_t) = 
      \begin{bmatrix}
        \cos \theta_t & -\sin \theta_t \\
        \sin \theta_t & \cos \theta_t
      \end{bmatrix}
    \]
    \Comment{Rotation matrix for angle \(\theta_t\) in 2D space}
    
    \State Compute Saturation operator:
    \[
      R_{zt} \leftarrow R_z \cdot \mathcal{R}(\theta_t)
    \]
    \Comment{Apply Saturation operator to the token embedding vector during propagation in training process}

    \State Apply operator:
    \[
      \tilde{e}_t \leftarrow e_t \cdot R_t
    \]
    \Comment{$\tilde{\text{emb}} := [\, \tilde{\mathbf{e}}_1 \quad \tilde{\mathbf{e}}_2 \quad \dots \quad \tilde{\mathbf{e}}_T \,]$}
    
  \end{algorithmic}
\end{algorithm}  

\vspace{1em}
\textbf{Usage:}

  \begin{algorithm}[H]
    \caption{Feature grounding}
    \begin{algorithmic}[1]

    \State Propagate through the model:
    \[
      emb, \tilde{\text{emb}} \leftarrow \text{model}(t)
    \]
    
    \State Compute reconstruction loss:
    \[
      \mathcal{L}_{\text{recon}} = \mathrm{MSE}(\tilde{\text{emb}}, X)
    \]
    
    \State Compute contrastive loss:
    \[
      \mathcal{L}_{\text{contrastive}} = \mathrm{contrastive}(emb_i, emb_j)
    \]
    \State Backpropagate loss via Embedding:
    \[
      \mathcal{L} = \mathcal{L}_{\text{recon}} + \mathcal{L}_{\text{contrastive}}
    \]

    \State After the pretraining phase, the embedding layer can be extracted from the model and used as a conventional embedding layer in downstream tasks.

  \end{algorithmic}
\end{algorithm}

\subsection{Pre-training Algorithm Diagram:}

\begin{figure}[H]
  \centering
  \includegraphics[width=0.8\textwidth]{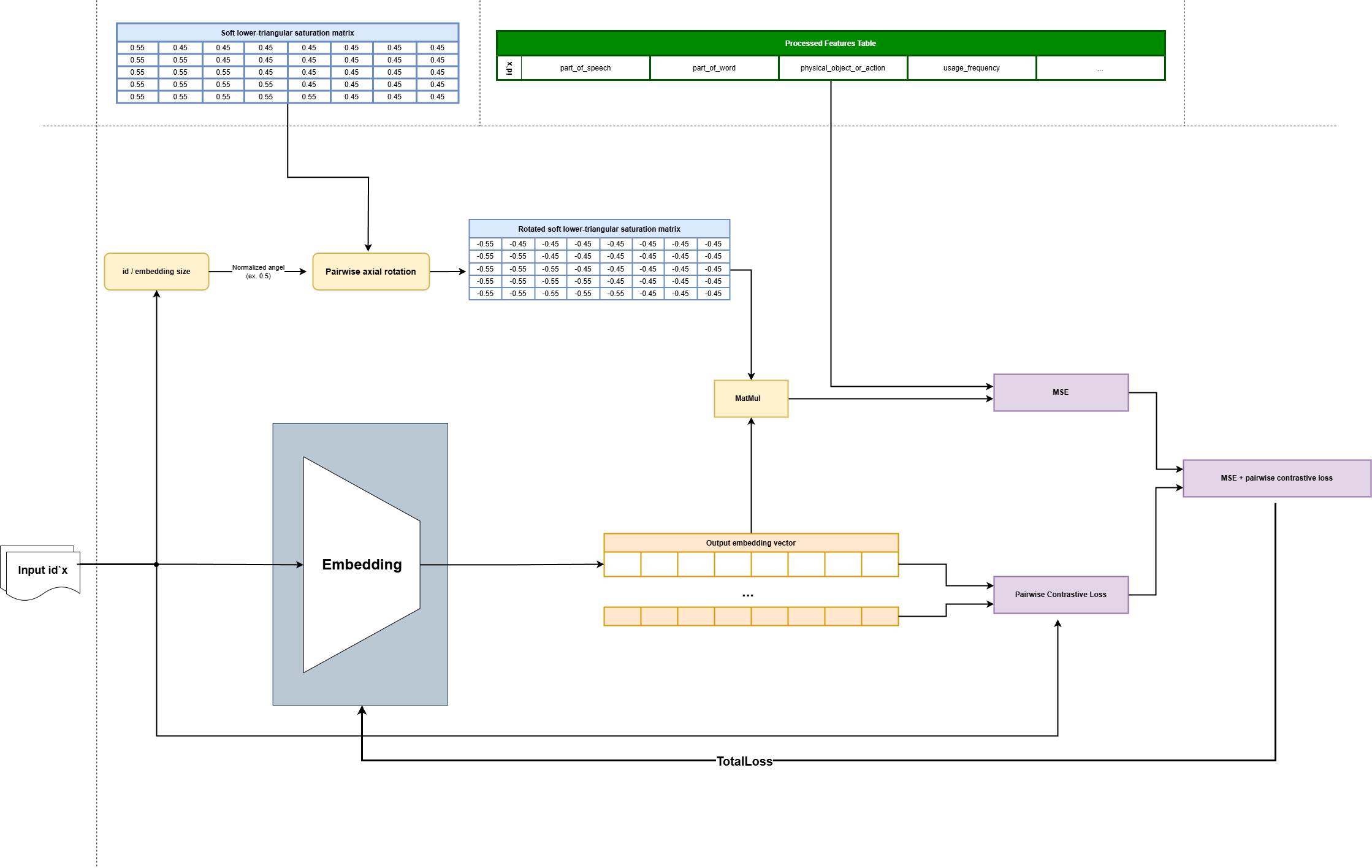}
  \caption{Description of the proposed method}
  \label{fig:training-algorithm}
\end{figure}

\subsection{Use Case Example:}

\begin{figure}[H]
  \centering
  \includegraphics[width=0.4
  \textwidth]{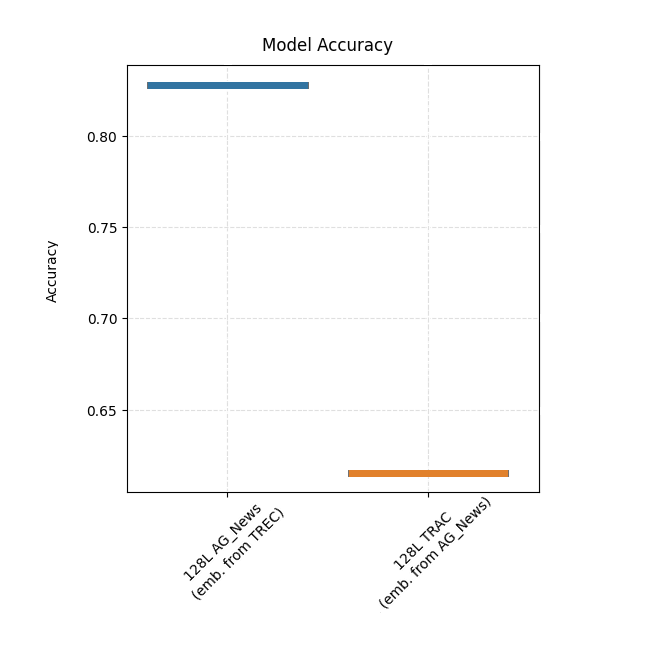}
  \caption{The TinyBERTClassifier with feature-grounded word-embeddings swapped after training on different datasets.}
  \label{fig:accuracy_128L_trec_ag_news}
\end{figure}

The Figure~\ref{fig:accuracy_128L_trec_ag_news} illustrates the performance of models after their word-embedding layer were swapped across two independently trained instances. 
Each models was trained separately on distinct datasets: one on the TREC dataset containing 50 target classes, and other on the AG News dataset containing 4 target classes. Evaluation was performed on the respective test sets corresponding to each training domain.
It can be observed that the model’s performance does not completely degrade. This observation suggests an alignment of semantic structures in latent representation across independently trained models. A similar experiment will be discussed in more detail in the experiment section.

\newpage
\section{Experiments}

\subsection{Setup:}

For the generation of grounding features, the tokenizer vocabulary was derived from the BERT tokenizer (provided by: \texttt{AutoTokenizer.from\_pretrained(``bert-base-uncased'')}).

Selected grouding features:
\begin{itemize}
  \item \textbf{part\_of\_speech}: \textit{noun}, \textit{verb}, \textit{adjective}, \textit{adverb}, \textit{preposition}, \textit{conjunction}, \textit{interjection}, \textit{pronoun}, \textit{numeral}, \textit{article}, \textit{particle}, \textit{modal\_verb}, \textit{auxiliary\_verb}, \textit{determiner}, \textit{none}
  \item \textbf{part\_of\_word}: \textit{prefix}, \textit{root}, \textit{suffix}, \textit{infix}, \textit{postfix}, \textit{circumfix}, \textit{none}
  \item \textbf{person}: \textit{first}, \textit{second}, \textit{third}, \textit{none}
  \item \textbf{connotation}: \textit{positive}, \textit{neutral}, \textit{negative}
  \item \textbf{physical\_object\_or\_action}: \textit{true}, \textit{false}
  \item \textbf{usage\_frequency}: \textit{s}, \textit{m}, \textit{l}, \textit{xl}
  \item \textbf{has\_many\_meanings}: \textit{true}, \textit{false}
  \item \textbf{can\_be\_used\_meaningfully\_on\_its\_own}: \textit{true}, \textit{false}
\end{itemize}

For the purpose of controlled experimentation and simplification, the word-embeddings were implemented using PyTorch's nn.Embedding module with default weight initialization method, corresponds to uniform initialization scaled by the embedding dimension.

Prior to feature-grounding, the vocabulary was filtered to exclude special tokens and pure character tokens (like single char).
This processed vocabulary was used to construct a token-to-index-to-feature mapping, which was used to supervise the training (grounding) of a word-embedding layer. 
This approach ensured compatibility with the BERT word-embedding layer and allowed for layer replacement.

\begin{figure}[H]
  \centering
  \includegraphics[width=0.8\textwidth]{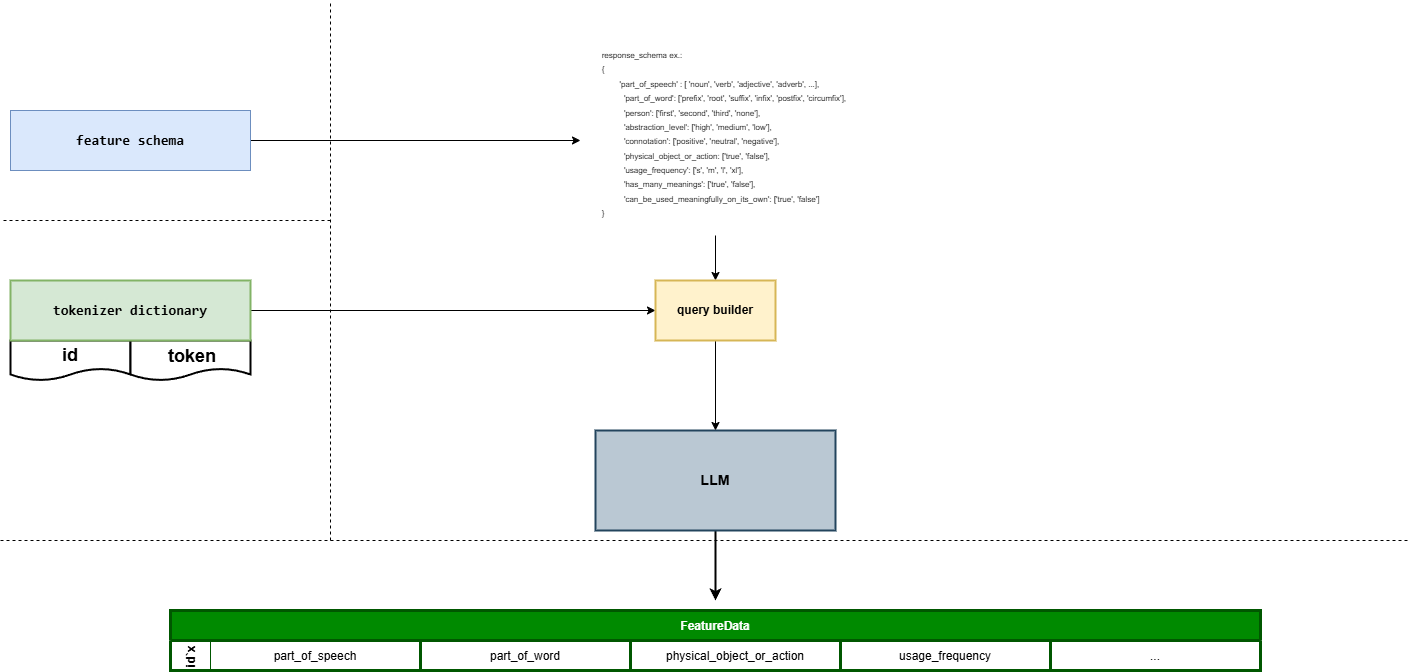}
  \caption{Extracting Grounding features}
  \label{fig:feature-distil}
\end{figure}

Feature-grounded word embedding was integrated with a lightweight classification model based on the \texttt{BERT} model and named \texttt{TinyBERTClassifier}.

The following datasets for training and evaluation \texttt{TinyBERTClassifier} were selected:
\begin{itemize}
  \item SST-2 Stanford Sentiment Treebank, binary classication (2 classes)
  \item AG News news topic classication with 4 distinct categories
  \item TREC question classication dataset containing 50 ne-grained target classes
\end{itemize}

  \vspace{1em}
\subsection{Grounding:}

\paragraph{Objectives:} 

\[
\mathcal{L} = 
\underbrace{\mathcal{L}_{\mathrm{MSE}}(\tilde{E}, X)}_{\text{Reconstruction Loss}} +   \underbrace{\mathcal{L}_{\mathrm{contrastive}}(\mathbf{e}_{t_i}, \mathbf{e}_{t_{j}})}_{\text{Min-max Distance Loss}}
\]

\begin{figure}[H]
  \centering
  \includegraphics[width=0.8\textwidth]{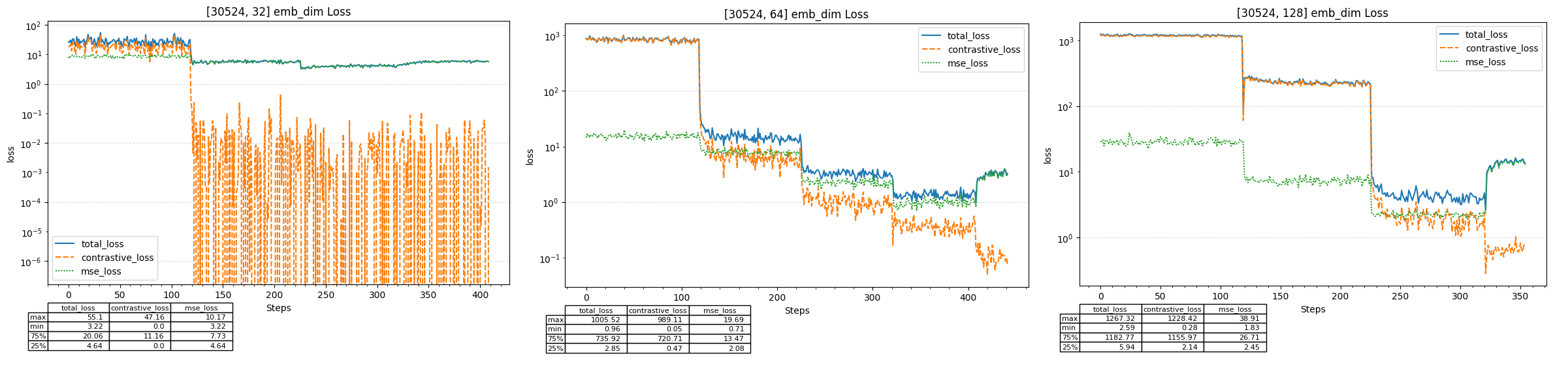}
  \caption{Losses during grounding}
  \label{fig:training-plot}
\end{figure}

\paragraph{Learning curves:} 
The Embedding loss dynamics during grounding over epochs are illustrated in Figure~\ref{fig:training-plot}.

\begin{figure}[H]
  \centering
  \includegraphics[width=0.8\textwidth]{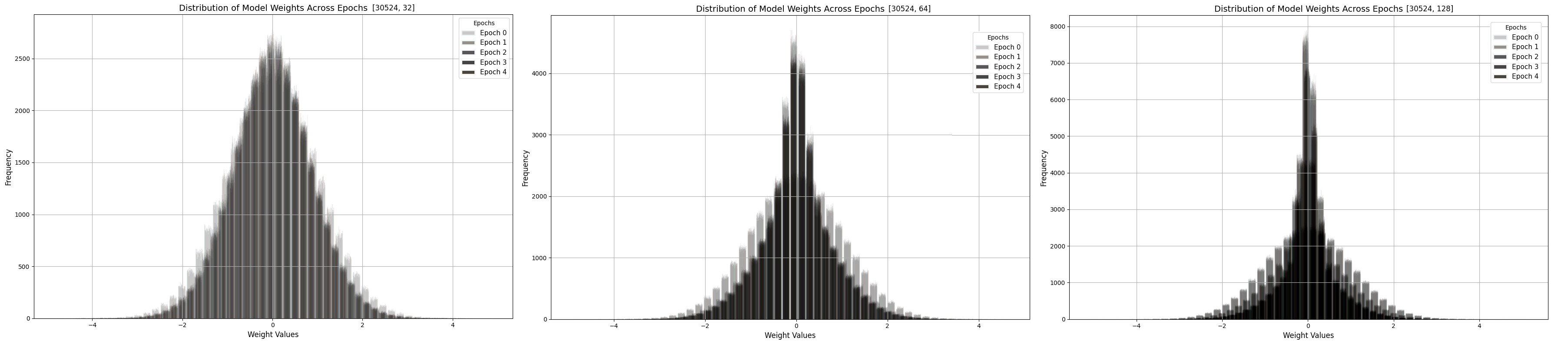}
  \caption{Weights distribution during training}
  \label{fig:weights-distribution}
\end{figure}

\paragraph{Weights behevior:} 
The Embedding weights distribution dynamics over epochs are illustrated in Figure~\ref{fig:weights-distribution}.

\vspace{1em}

\vspace{1em}
\subsection{Usage:}

After the embeddings was grounded, they were extracted from the training model and inserted into a \texttt{TinyBERTClassifier}, like a word-embeddings. 
Then, the entire \texttt{TinyBERTClassifier} was trained and evaluated.

\vspace{2em}
For each embedding size, two independent networks were trained:
\begin{itemize}
  \item the base variant, denoted as \texttt{size},
  \item and an extended variant, denoted as \texttt{size\_L}, which was trained longer
\end{itemize}
Both models will not fit perfectly, but for experiment purposes, the process was limited by time.

\vspace{2em}

After training, the word-embeddings from the two variants (\texttt{size} and \texttt{size\_L}) were swapped across the models, and the resulting performance was evaluated again.

The core idea of the experiments is prove that the provided mechanism enables the interchange of model-modules between different models even those trained on distinct corpora without significantly degrading their performance. 

This observation proves that model`s internal structure semantically similar and consistent across independently trained networks, facilitating transferability and better interoperability.

Surprisingly, the performance degradation after embedding brutally was swapped is minor and in several cases performance even improves.

\begin{figure}[H]
  \centering
  \includegraphics[width=0.8\textwidth]{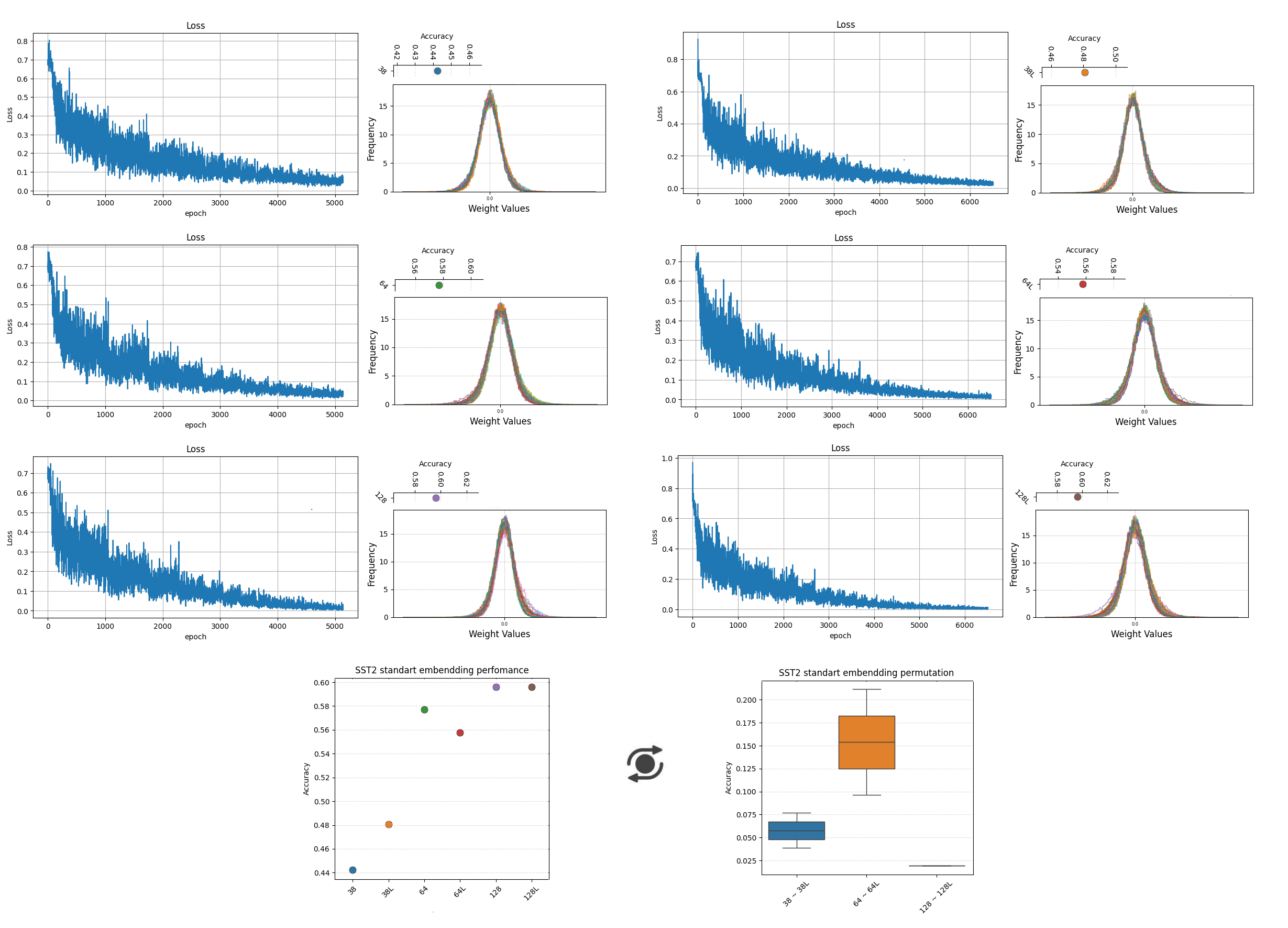}
  \caption{TinyBERTClassifier losses with standard embeddingSST2 dataset}
  \label{fig:training-loss}
\end{figure}

\begin{figure}[H]
  \centering
  \includegraphics[width=0.8\textwidth]{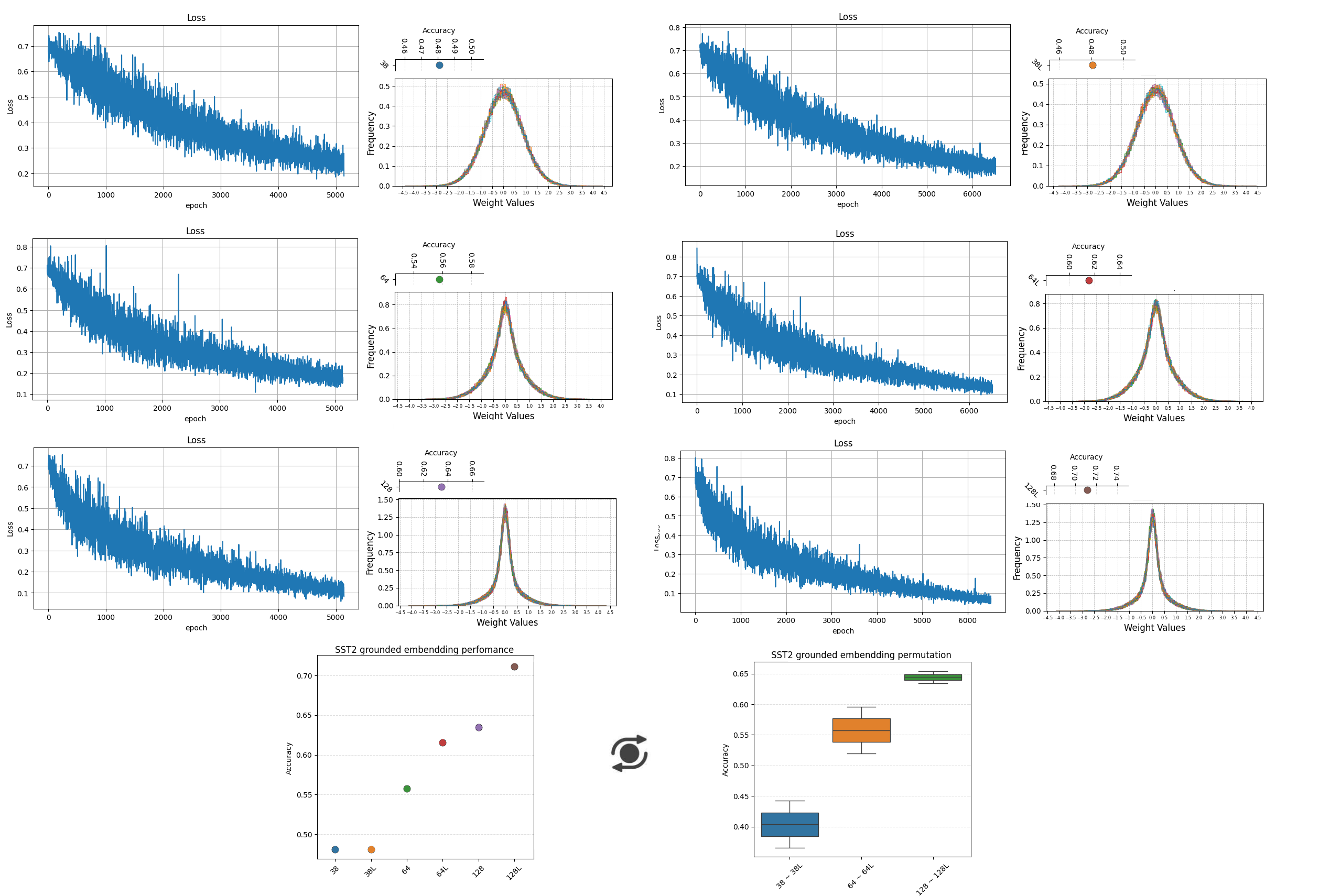}
  \caption{TinyBERTClassifier losses with feature-grounding rotated embedding SST2 dataset}
  \label{fig:training-loss}
\end{figure}

\begin{figure}[H]
  \centering
  \includegraphics[width=0.8\textwidth]{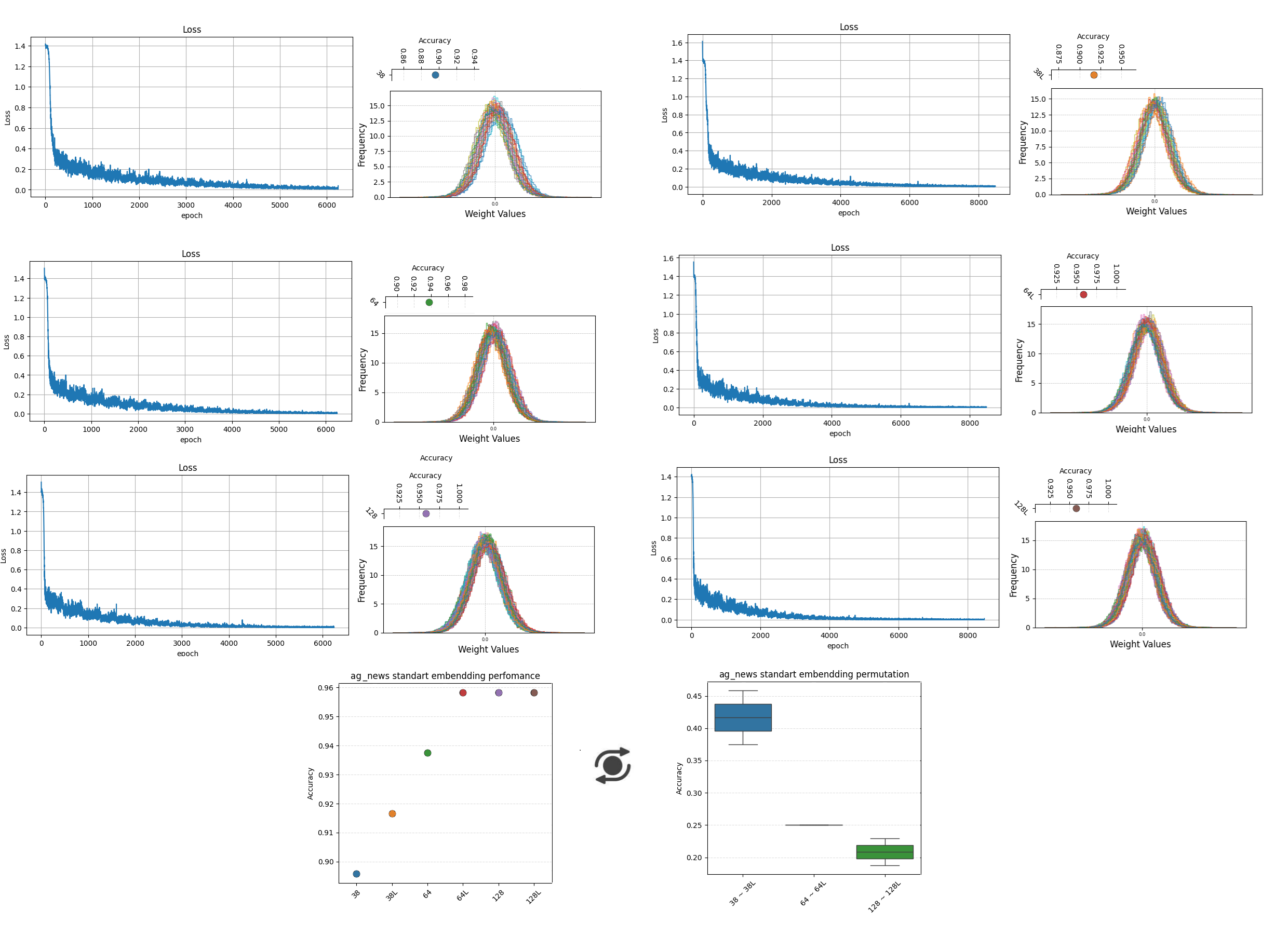}
  \caption{TinyBERTClassifier losses with standard embeddingAG News dataset}
  \label{fig:training-loss}
\end{figure}

\begin{figure}[H]
  \centering
  \includegraphics[width=0.8\textwidth]{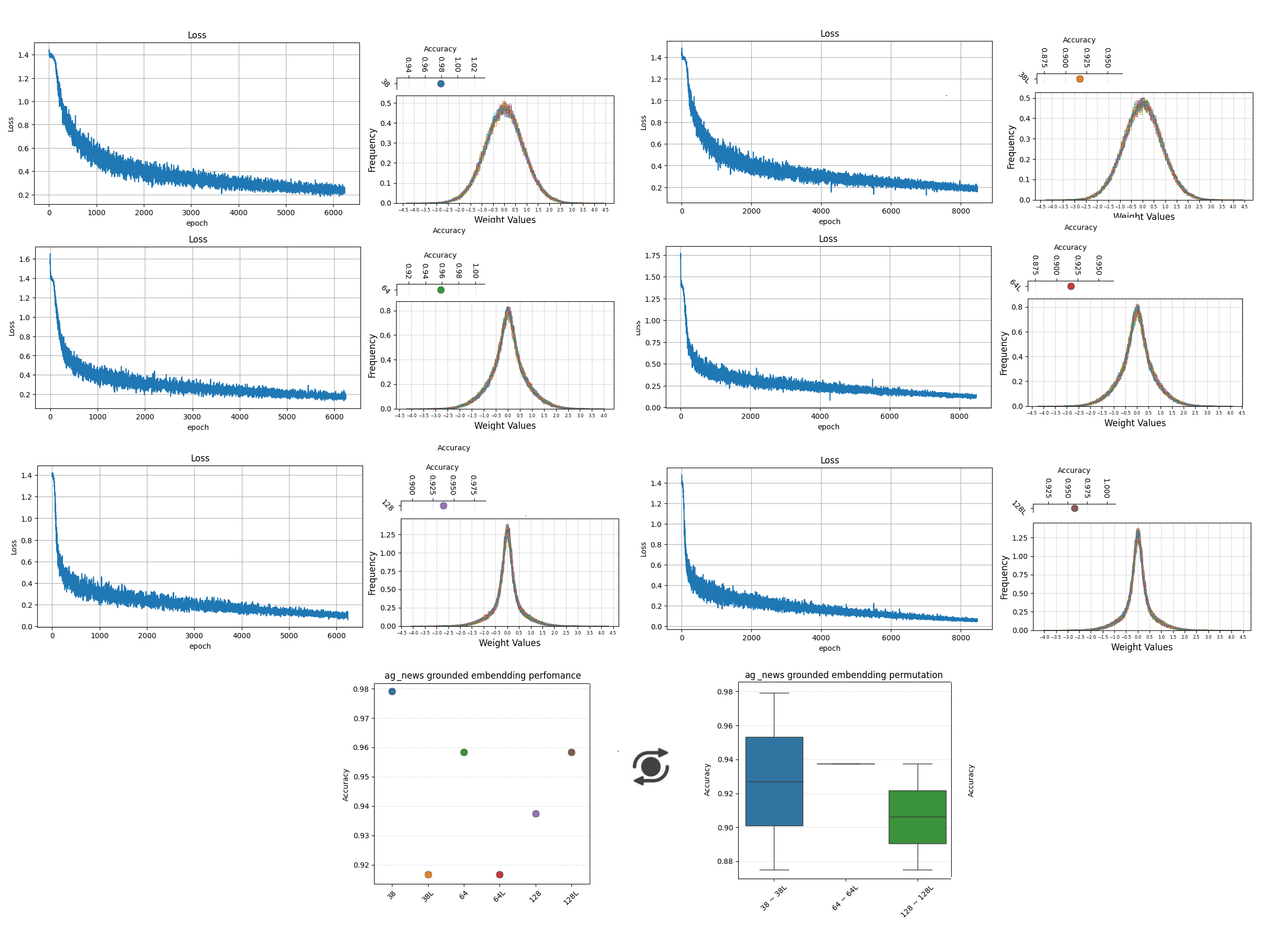}
  \caption{TinyBERTClassifier losses with feature-grounding rotated embedding AG News dataset}
  \label{fig:training-loss}
\end{figure}

\begin{figure}[H]
  \centering
  \includegraphics[width=0.8\textwidth]{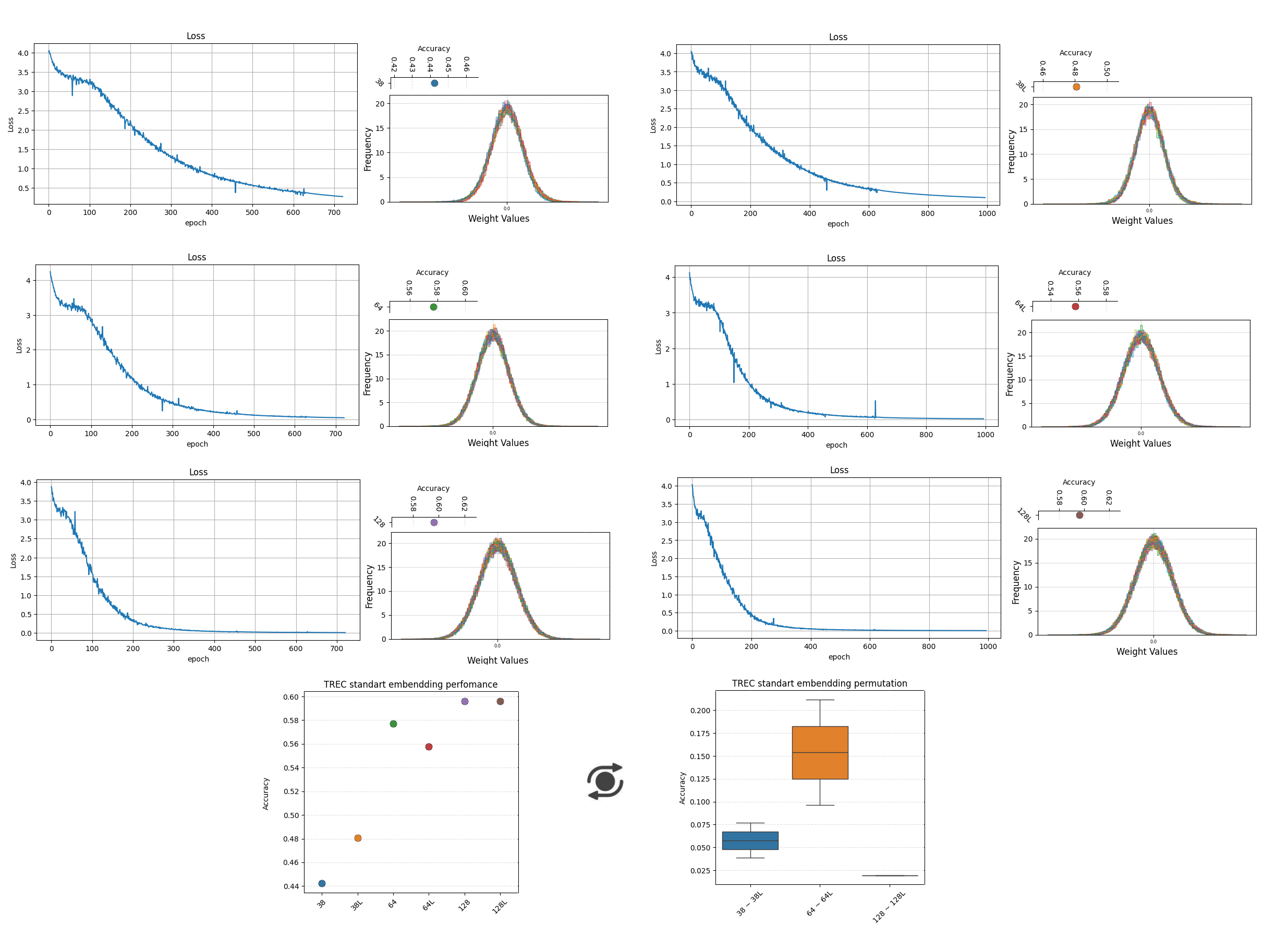}
  \caption{TinyBERTClassifier losses with standard embeddingTREC dataset}
  \label{fig:training-loss}
\end{figure}

\begin{figure}[H]
  \centering
  \includegraphics[width=0.8\textwidth]{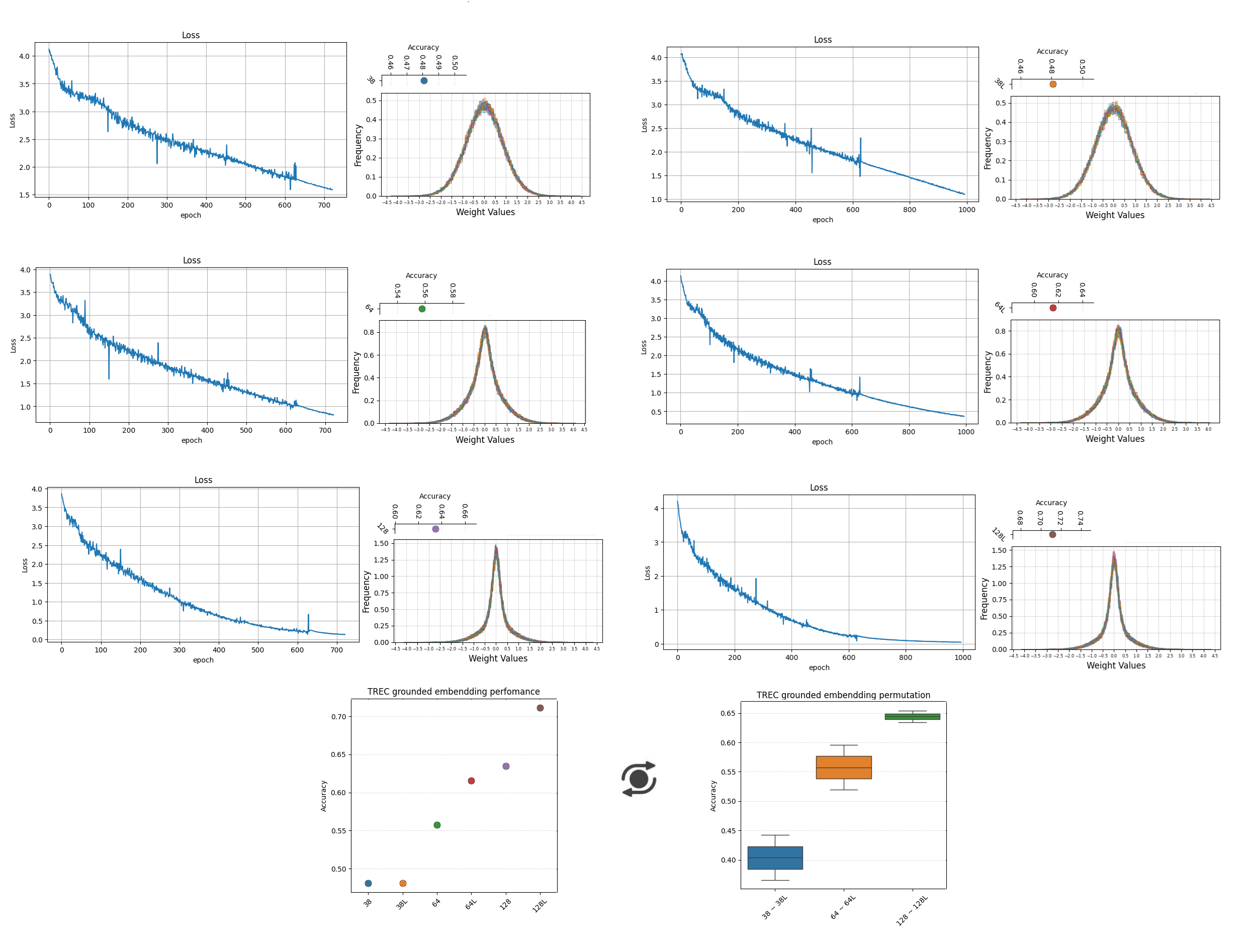}
  \caption{TinyBERTClassifier losses with feature-grounding rotated embedding TREC dataset}
  \label{fig:training-loss}
\end{figure}

\begin{figure}[H]
  \centering
  \includegraphics[width=0.8\textwidth]{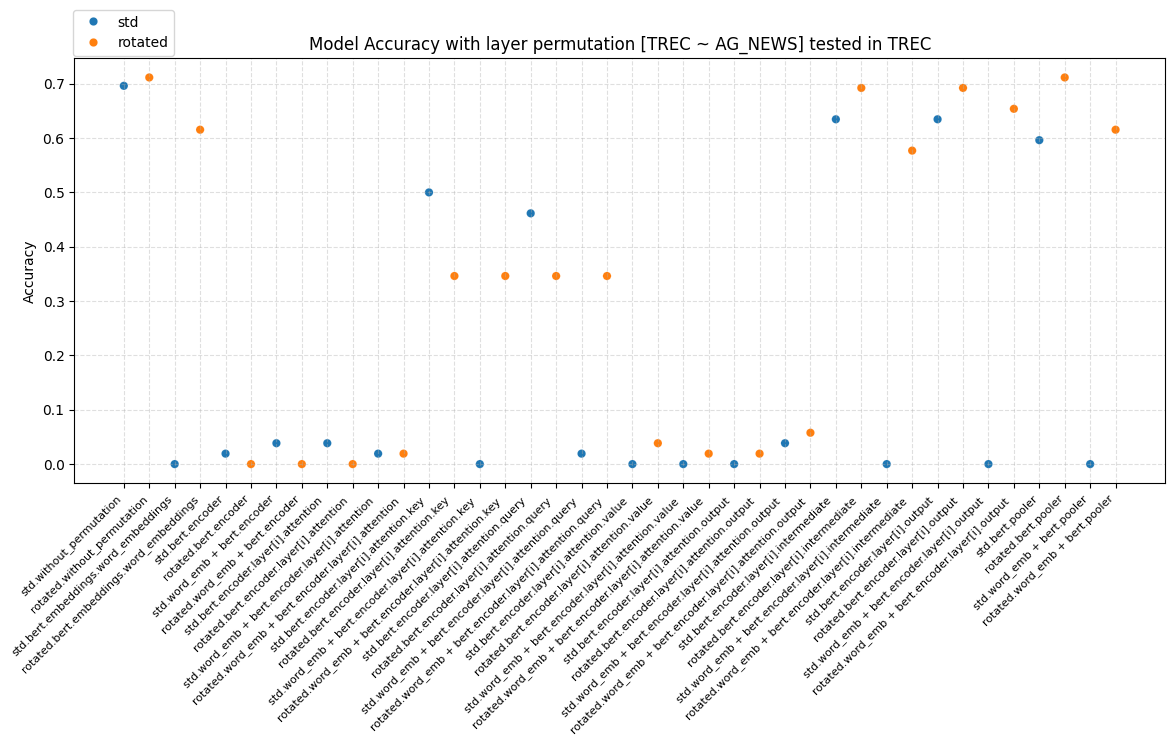}
  \caption{TinyBERTClassifier performance after layers was swapped}
  \label{fig:layers_permutation}
\end{figure}

The Figure~\ref{fig:layers_permutation} illustrates the performance of models after their layers were swapped across each other.
std - model standardBERT classifier.
rotated - model with word-Embedding with grounding.
For each test there was used model pairs:
\begin{itemize}
  \item rotated.TinyBERTClassifier(trec) and rotated.TinyBERTClassifier(learned in ag\_news)
  \item std.TinyBERTClassifier(learned in trec) and std.TinyBERTClassifier(learned in ag\_news)
\end{itemize}
All models were evaluated on the TREC dataset.
This observation highlights where alignment between internal representations can be observed in this specific experimental setup.

\section{Discussion}

The presented results confirm that the proposed method consistently induces an interpretable internal structure within the hidden space of the model. 
This structure exhibits a degree of predictability and regularity, which enables the replacement and interchange of trained modules across different model instances.
Such modularity, achieved without extensive re-training, suggests the possibility of constructing larger systems from reusable components.
This could be particularly beneficial in scenarios involving user-adaptive inference, distributed learning or controlled access to model internals.

Moreover, the observed effect raises further questions regarding the internal geometry of hidden spaces. 
In particular, the notion of a \emph{rotated projector} or \emph{saturation operator} may offer an alternative mechanism for encoding a greater volume of information within fixed-dimensional representations.
This could potentially reduce the curse of dimensionality by promoting disentanglement and improved semantic coherence in learned features.

Nevertheless, the present study is limited to relatively simple architectures and controlled training conditions. 
Further insights require investigating how the model performs when initialized with different weights range prior to grounding. 
Additionally, it is crucial to assess the model's behavior when extended features are incorporated into the grounded layers.
The extent to which the observed structural alignment generalizes to larger-scale models or heterogeneous data domains remains an open question.

\section{Conclusion}

The conducted experiment demonstrates that the proposed method enables intentional shaping of a model's loss-space structure through feature-grounding guided by known priors.
By enforcing a predictable internal representation, this approach facilitates hidden-space unification across models with similar architectures, thereby enhancing the interpretability and modularity of neural systems.
Such structural regularity not only contributes to a deeper understanding of model behavior, but also opens the possibility of constructing specialized models incorporating variational knowledge (e.g., user-level permissions), where model parts where grounded with distinct knowledge.

These findings highlight the potential of feature-grounding as a practical tool for developing modular, interpretable and reusable neural components, especially in contexts requiring flexibility, security or distributed model governance.

\section*{Acknowledgements}

I would like to acknowledge the use of the following datasets: **AG News** \cite{ag_news}, **SST-2** \cite{sst2}, and **TREC** \cite{trec}. These datasets were instrumental in the development and evaluation of the models.

Additionally, we would like to thank Hugging Face for providing the **BERT** model \cite{bert} with attention implementation\cite{vaswani2017attention}, which was utilized for experiments. The implementation of the BERT model from Hugging Face’s Transformers library has significantly facilitated the rapid development of this work.

\bibliographystyle{plain}
\bibliography{references}

\end{document}